\documentclass[a4paper,fleqn]{cas-dc}

\usepackage[numbers]{natbib}

\def\tsc#1{\csdef{#1}{\textsc{\lowercase{#1}}\xspace}}
\tsc{WGM}
\tsc{QE}
\tsc{EP}
\tsc{PMS}
\tsc{BEC}
\tsc{DE}

\begin{document}
\let\WriteBookmarks\relax
\def\floatpagepagefraction{1}
\def\textpagefraction{.001}

\shorttitle{LEGO: Self-Supervised Representation Learning for Scene Text Images}

\shortauthors{Yujin Ren et~al.}

\title[mode = title]{LEGO: Local Explicit and Global Order-aware Self-Supervised Representation Learning for Scene Text Images}                   

%

\author[1]{Yujin Ren}[type=editor,
                      style=chinese,
                      auid=000,bioid=1]
\author[1]{Jiaxin Zhang}[type=editor,
                         style=chinese,
                         auid=000,bioid=1]
\author[1]{Lianwen Jin\corref{cor1}}[style=chinese]





\credit{Conceptualization of this study, Methodology, Software}

\affiliation[1]{organization={South China University of Technology},
    city={Guangzhou},
    country={China}}

\credit{Data curation, Writing - Original draft preparation}

\cortext[cor1]{Corresponding author}

\begin{abstract}
In recent years, significant progress has been made in scene text recognition by data-driven methods. However, due to the scarcity of annotated real-world data, the training of these methods predominantly relies on synthetic data. The distribution gap between synthetic and real data constrains the further performance improvement of these methods in real-world applications. To tackle this problem, a highly promising approach is to utilize massive amounts of unlabeled real data for self-supervised training, which has been widely proven effective in many NLP and CV tasks. Nevertheless, generic self-supervised methods are unsuitable for scene text images due to their sequential nature. To address this issue, we propose a Local Explicit and Global Order-aware self-supervised representation learning method (LEGO) that accounts for the characteristics of scene text images. Inspired by the human cognitive process of learning words, which involves spelling, reading, and writing, we propose three novel pre-text tasks for LEGO to model sequential, semantic, and structural features, respectively. The entire pre-training process is optimized by using a consistent Text Knowledge Codebook. Extensive experiments validate that LEGO outperforms previous scene text self-supervised methods. The recognizer incorporated with our pre-trained model achieves superior or comparable performance compared to state-of-the-art scene text recognition methods on six benchmarks. Furthermore, we demonstrate that LEGO can achieve superior performance in other text-related tasks.

\end{abstract}


\maketitle

\section{Introduction}

Over the last decade, data-driven deep learning methods have achieved considerable advancement in several computer vision tasks, including scene text recognition. Among these methods, supervised ones necessitate large-scale labeled data for training purposes, otherwise, models are prone to performance degradation and generalization problems. The most efficient way to acquire large-scale labeled data involves employing data synthesis techniques~\cite{gupta2016synthetic, jaderberg2014synthetic, long2020unrealtext}, as manual annotation can be exceedingly time-consuming and labor-intensive. However, the reliance on synthetic data limits further progress due to the inherent domain gap between synthetic and real data. 
To address this challenge, self-supervised learning (SSL) has emerged as a promising alternative. This approach capitalizes on vast amounts of unlabeled real-world data for training by extracting supervised information directly from input images. The efficacy of SSL has been widely demonstrated~\cite{chen2020simple, he2020momentum, bao2021beit, he2022mae, xie2022simmim}, showcasing its potential as a viable approach for advancing various computer vision tasks, including scene text recognition.

Pioneering SSL~\cite{chen2020simple, he2020momentum} in the field of computer vision primarily targets natural scene images. Applying these methods directly to text images results in significant performance degradation due to certain inherent characteristics specific to text, such as the sequential nature. While many efforts~\cite{aberdam2021sequence, liu2022perceiving, luo2022siman, yang2022reading} have been made to tailor these methods for text images, they still overlook other crucial attributes of them. In the following, we delve into three critical characteristics of text and elucidate how they contribute to the suboptimal performance of the above methods:

\begin{figure}[t]
\centering
\small
\includegraphics[scale=.5]{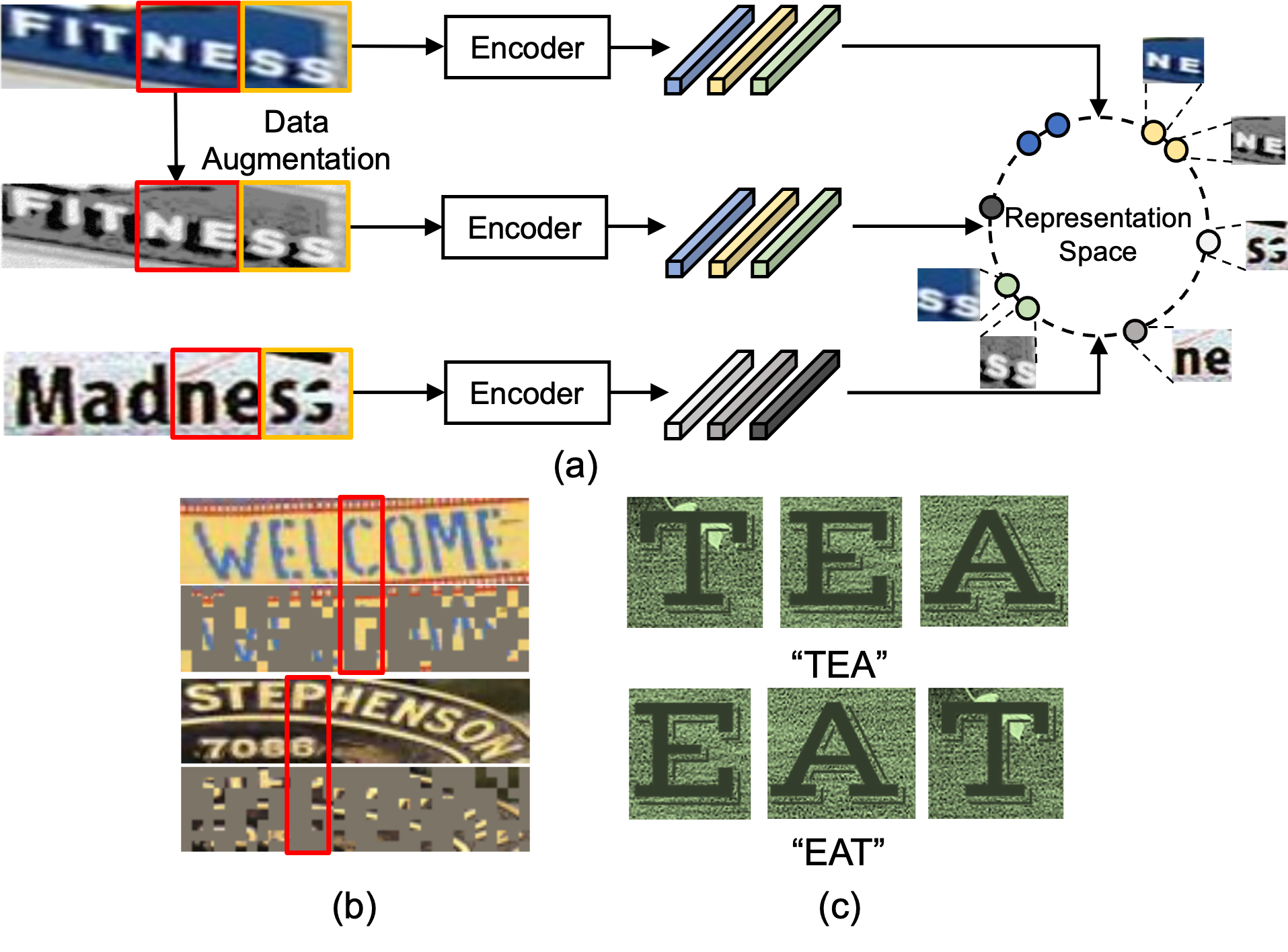}
\caption{Problems encountered when performing self-supervised learning (SSL) on scene text images. (a) Ambiguity brought by rough sample division strategy of contrastive learning, (b) Indeterminacy due to invalid reconstruction targets caused by random masking, and (c) Sequentiality, a natural property of text that the same letters in different orders can form words with different meanings.}
\label{fig:motivation}
\end{figure}

\textbf{Hierarchy:} Text images exhibit hierarchy, with the most granular units being individual characters. Taking this into consideration, methods like SeqCLR~\cite{aberdam2021sequence, liu2022perceiving, yang2022reading} have improved upon standard contrastive learning approaches~\cite{chen2020simple,he2020momentum} (based on "whole-image" as the basic unit) by adapting them to operate on image slices (sub-word), thus better suiting the hierarchical nature of text images. However, due to fewer characters in these slice units, the diversity is significantly reduced compared to the whole image. This greatly increases the probability of positive and negative samples containing the same content in contrastive learning (as shown in Fig.~\ref{fig:motivation} (a)), leading to ambiguity in model learning.

\textbf{High information density:} Common generative SSL methods~\cite{he2022mae,xie2022simmim} rely on Mask Image Modeling (MIM), where image patches are masked at a considerable ratio, and the network is tasked with reconstructing the masked content from the remaining visible portion. While some existing text image SSL methods~\cite{yang2022reading, lyu2022maskocr} directly follow this way, we contend that it is less appropriate. Text images, as opposed to general object images, tend to possess higher information density. Randomly masking large portions of text images risks losing entire characters and introducing excessive background noise, which may not offer sufficient effective structural cues for guiding MIM. As illustrated in Fig.~\ref{fig:motivation} (b), instructing the network to reconstruct such heavily masked images presents significant challenges and may not be practical.

\textbf{Sequentiality:} Unlike general object images, wherein changes in the positions of internal elements may not significantly alter semantic information, such adjustments in text images can result in entirely distinct meanings. As depicted in Fig.~\ref{fig:motivation} (c), the rearrangement of the letters in "TEA" to form "EAT" maintains three identical characters but conveys distinctly different meanings. This fundamental characteristic has often been overlooked in previous approaches. Modeling the connectivity between different portions of text images is advantageous for high-level downstream tasks like text recognition.

Considering these features of text images, we propose an innovative self-supervised learning method, called Local Explicit and Global Order-aware (LEGO). It integrates three key pretext tasks: Selective Individual Discrimination (SID), Enhanced Mask Image Modeling (MIM), and Random Text Rearrangement (RTR). All these tasks are facilitated with the assistance of our novel Text Knowledge Codebook which is a text-tailored discrete quantizer based on VQVAE~\cite{van2017neural}. It aggregates features from slices with similar structures and semantics, replacing them with identical latent space vectors. More specifically, within the SID task, the Text Knowledge Codebook is utilized to filter out incorrect negative samples caused by the hierarchical nature of the text. In the MIM task, the Text Knowledge Codebook provides additional information to alleviate significant information loss. Finally, in the RTR task, it assists in providing ground truth for sequence order.

LEGO's pretraining process simultaneously incorporates these three pretext tasks, mimicking the cognitive process of human learning to read, write, and spell. Specifically, the SID task employs contrastive learning to distinguish semantic information among different characters, resembling the step of reading; the MIM task is utilized to learn the structural information of text, akin to the process of writing; the RTR task implicitly captures linguistic information contained within words, inspired by the spelling process. These tasks collectively enhance the model's representation capability. Moreover, our Text Knowledge Codebook encapsulates consistent text information, enabling it to simultaneously assist in the aforementioned three pretext tasks, akin to humans referencing a dictionary when learning words.

To summarize, our main contributions are three-fold:
\begin{enumerate}
\item Taking into account the characteristics of text image, we propose a novel SSL method, LEGO, which integrates discriminative, generative, and sequential pretext tasks (i.e. SID, MIM, and RTR). These tasks respectively mimic the cognitive processes of reading, writing, and spelling, collectively enhancing the model's representation capability for text images.

\item A novel Text Knowledge Codebook is designed to assist these pretext tasks by filtering out incorrect negative samples in SID, alleviating significant information
loss in MIM, and providing ground truth for sequence order in RTR.

\item Experiments on downstream tasks demonstrate that the model pre-trained by our approach can improve the performance of scene text recognition and scene text super-resolution.
\end{enumerate}

\section{Related Work}

\subsection{Scene Text Recognition}
Scene text recognition (STR) has attracted considerable academic attention due to its wide application value. In the deep learning era, many methods have converted STR into a sequence-to-sequence task and made significant breakthroughs through the encoder-decoder structure. According to the category of decoder, they can be divided into CTC-, Attention- and Transformer-based three types. CRNN~\cite{shi2016crnn} firstly combines CNN and RNN to extract visual features and model features sequence, then through a CTC~\cite{graves2006ctc} decoder to maximize posterior probability by location in all paths for the sequence prediction. However, CTC-based methods~\cite{wang2017gated, liu2018connectionist, hu2020gtc} assume the text is horizontal so that they cannot handle the text of irregular shapes such as curvature and perspective distortion, which are the more common forms that appear in scene text images. To this end, attention-based~\cite{luo2019moran, shi2018aster} methods were proposed because of their natural advantages in spatial localization. The variants of attention-based decoders~\cite{li2019sar, cheng2023lister} were popular in previous text recognizers. Recently, with Transformer sweeping across plenty of fields, STR methods have evolved once again~\cite{tao2021trig, tan2022pure}. Due to the compatibility of the Transformer architecture, the language model can be integrated into the recognizer to rectify mistakes caused by uncleared or incomplete vision, resulting in impressive performance~\cite{fang2021read, wang2021two, wang2022multi}. However, all of the above methods are limited by the scale and quality of the training data.

\subsection{Self-Supervised Pre-training for Text Images}

To alleviate the data hunger issue of text recognizers based on deep learning, an increasing number of self-supervised methods for text have been proposed recently. Existing state-of-the-art methods can be summarized into three main categories: contrastive learning scheme~\cite{aberdam2021sequence,liu2022perceiving}, generative learning scheme~\cite{luo2022siman, lyu2022maskocr} and a combination of the two~\cite{yang2022reading}. SeqCLR~\cite{aberdam2021sequence} firstly introduced contrastive learning into text and treated an image as a sequence of frames to satisfy the sequence-to-sequence format of text recognition. PerSec~\cite{liu2022perceiving} learned latent representations from low-level stroke and high-level semantic contextual spaces simultaneously by contrastive learning. For the generative learning scheme, SimAN~\cite{luo2022siman} studied the unique properties of scene text and used an image reconstruction pretext task. DiG~\cite{yang2022reading} integrates contrastive learning and masked image modeling into a unified model, taking advantage of the benefits of both.

Compared with the above-mentioned methods, we focus on alleviating the local ambiguity and the global weak ties that occur during pre-training. We propose a novel representation learning scheme equipped with the Text Knowledge Codebook, which introduces prior into all the processes of contrastive, generative, and sequential learning.

\subsection{Discrete Visual Representation}

Latent visual representation is a key open issue in the field of image generation and has shown wide application potential. VQVAE~\cite{van2017neural}, as one of the classic methods, learned discrete representations of images by utilizing CNN to model their distribution auto-regressively. To synthesize images with high resolution, VQGAN~\cite{esser2021taming} proposed a context-rich discrete visual representation, whose composition is subsequently modeled with an auto-regressive transformer architecture. When combined with a language-image pre-trained model, text-guided images~\cite{ding2021cogview,yu2022scaling} and videos~\cite{menapace2021playable,park2021vid} generation can be realized. Besides, discrete visual representation can provide an image token as a reconstruction target, enabling visual self-supervised learning~\cite{bao2021beit} to perform similar tasks to the Mask Language Model (MLM) in NLP self-supervised learning~\cite{devlin2018bert}. After this pre-training process, the model implements the ability to distinguish context between different regions. In this paper, we tokenize the structure and semantic information of each text image patch through discrete visual representation, to obtain unified text prior knowledge.

\section{Methodology}

We adhere to the standard self-supervised learning process, where the image encoder obtained through SSL pre-training is embedded into decoders specific to certain tasks for fine-tuning. The ultimate purpose of LEGO is to provide such a self-supervised pre-trained image encoder.

\begin{figure}[htb]
\centering
\includegraphics[scale=.42]{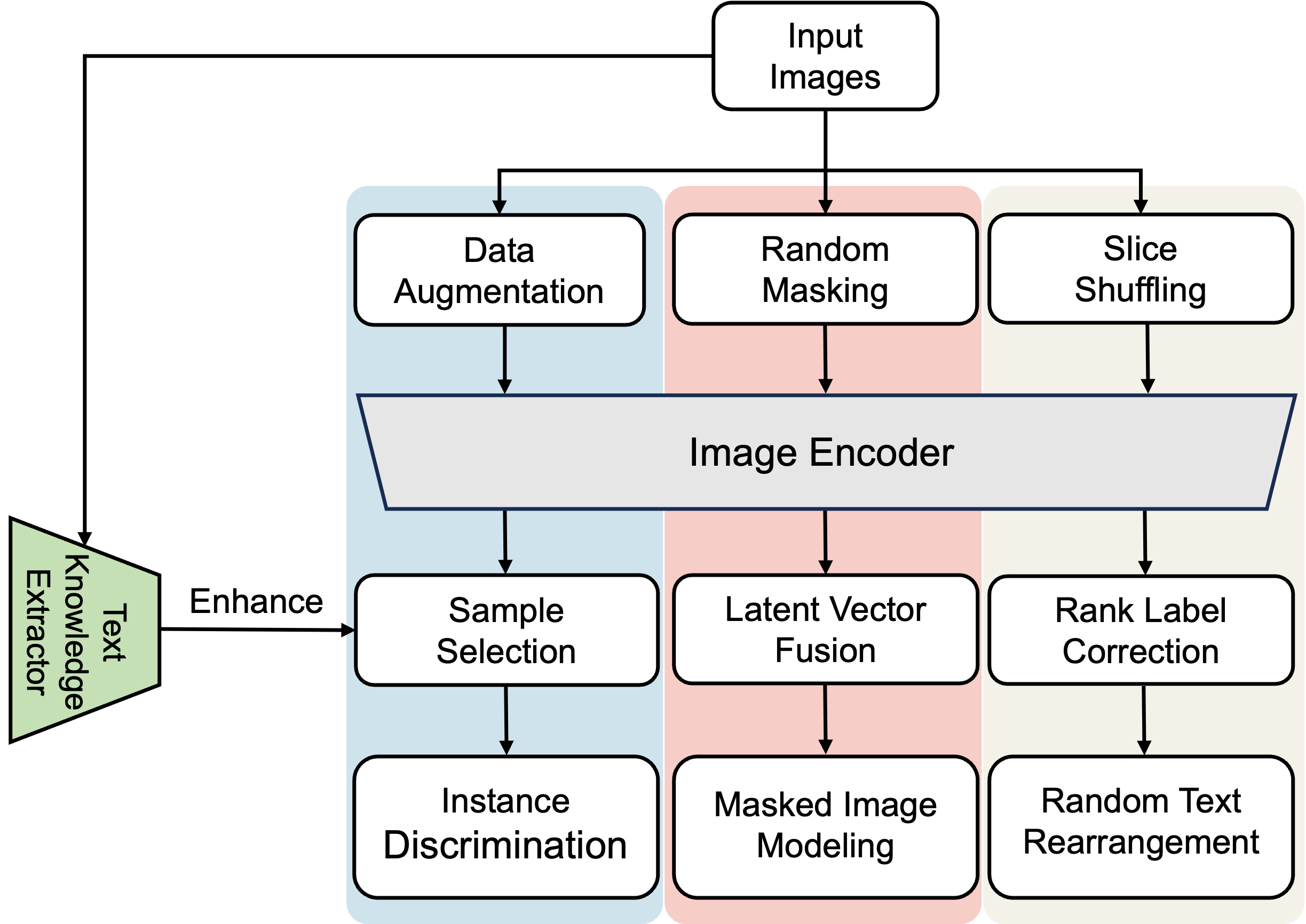}
\caption{The pipeline of LEGO. The T-Encoder depicted on the left is derived from T-VQVAE training. Before SSL pre-training, we feed input images into the T-Encoder to obtain the unified Text Knowledge Codebook. After three pretext tasks (shown on the right) designed for scene text SSL, the pre-trained ViT encoder can be migrated to downstream tasks.}
\label{fig2}
\end{figure}

Fig.~\ref{fig2} provides an overview of the proposed LEGO, which comprises three novel pretext tasks guided by consistent text information from the Text Knowledge Codebook. In summary, this codebook is obtained by our T-VQVAE (described in the following Section 3.1), whose parameters are frozen once its training is completed. During the self-supervised pre-training process, the Text Knowledge Codebook serves as an ``image tokenizer" to convert raw pixels into latent vectors, each of which corresponds to a discrete token index. Such codebook maps text patches with similar structure and context to the same index, enabling the extraction of consistent prior knowledge from the scene text images. With the aid of that, the model is designed to perform discriminative, generative, and sequential pretext tasks, which respectively mimic human reading, writing, and spelling. Three pre-processing operations are applied for these three pretext tasks respectively. In this manner, the representation quality can be progressively improved via SSL.

In the following, we first introduce the Text Knowledge Codebook in Section~\ref{sec:Text Knowledge Codebook} Subsequently, the three pre-text tasks are elaborated in detail in Section ~\ref{sec:Selective Individual Discrimination}, Section~\ref{sec:Modified Masked Image Modeling}, and Section~\ref{sec:Random-ordered Text Rearrangement}, respectively. Lastly, Section 3.5 presents the final loss function used for training our model.

\subsection{Text Knowledge Codebook}
\label{sec:Text Knowledge Codebook}
\begin{figure*}[htb]
\centering
\includegraphics[scale=.6]{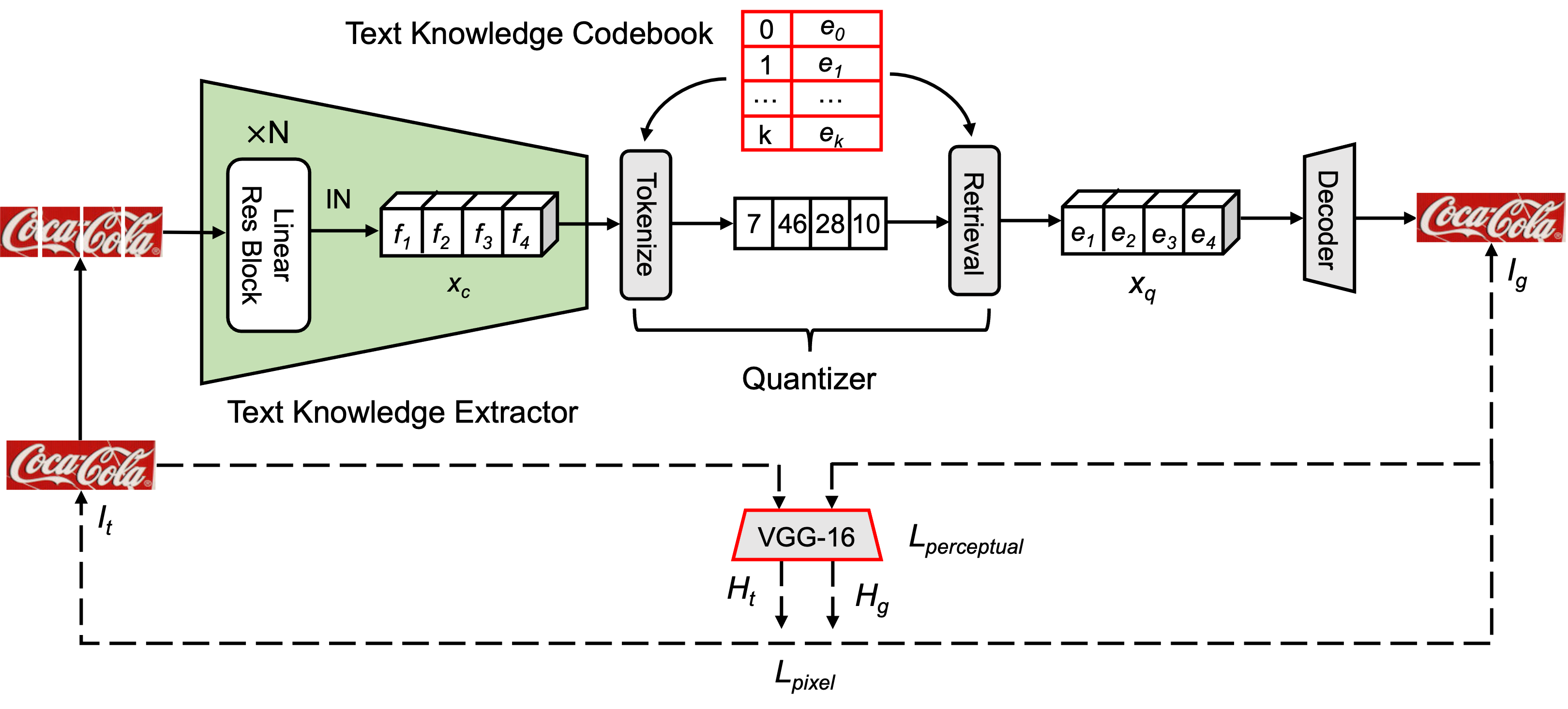}
\caption{The architecture of T-VQVAE. The quantizer aggregates de-styled text features in the latent vector space to generate the Text Knowledge Codebook through tokenization and retrieval. An officially trained VGG-16 is employed to calculate the perceptual loss.}
\label{fig3}
\end{figure*}

Inspired by the vector quantized auto-encoders~\cite{van2017neural,esser2021taming} used in text-to-image generation methods~\cite{ramesh2021zero,yu2022scaling}, we construct a text-tailored T-VQVAE. The biggest difference between the classic VQVAE and our approach is that we focus on the semantic content of text images, rather than the stylistic information coupled to them. As shown in Fig.~\ref{fig3}, T-VQVAE contains three major components: a T-Encoder, a Quantizer, and a Decoder. Continuous images can be encoded into learnable embeddings and discrete indexes, constituting a Text Knowledge Codebook that plays an important role in subsequent SSL pretext tasks.

Unlike strong semantic continuity between adjacent regions in general scenarios, the context information may change abruptly in text. The conventional CNN-based encoder in VQVAE inevitably introduces feature interaction when the kernel window slides. Therefore, we propose a more suitable T-encoder. Firstly, the input image $I_t$ is divided into $\frac{H}{r_1}\times \frac{W}{r_2}$ non-overlapping patches using convolutional layers with equal kernel size and stride, where $(r_1,r_2)$ is the spatial size of patches. For the sake of avoiding mutual disturbance between neighbors, we use several linear layers with residual connections to map high-dimensional visual patches into feature vectors $x_f\in\mathbb{R}^{\frac{H}{r_1}\times\frac{W}{r_2}\times C}$, where $C$ is the channel dimension of features. The content and style information of text images are coupled tightly, but we prefer T-Encoder to focus more on encoding text contents rather than stylistic differences between patches. According to the findings confirmed in previous studies~\cite{karras2019style}, the statistics of feature maps (i.e. mean and variance) can represent style. Hence, we perform instance normalization (IN) on $x_f$ to remove style and obtain text content vectors $x_c\in\mathbb{R}^{\frac{H}{r_1}\times\frac{W}{r_2}\times D}$:

\begin{center}
\begin{equation}
    x_c=IN(x_f)={\frac{x_f-\mu(x_f)}{\sigma(x_f)}}~,
\end{equation}
\end{center}
where $\mu(\cdot)$ and $\sigma(\cdot)$ represent the calculation of mean and standard deviation, respectively.

Then the Quantizer converts each vector of $x_c$ into the nearest neighbor embedding in a latent space $\{e_0,\ldots,e_{N-1}\} \in \mathbb{R}^D$, where each embedding corresponds to a discrete index $z\in\{0,\ldots,N-1\}$ ($N$ is set to 512 by default):
\begin{equation}
x_q=Quantizer(x_c)=e_k \in\mathbb{R}^{\frac{H}{r_1}\times\frac{W}{r_2}\times D}.
\end{equation}
Here $k=\mathop{argmin}\limits_{0\le i\le N-1} \|x_c-e_i\|_2$. 
Because the quantization process is non-differentiable, we adopt the straight-through estimator~\cite{van2017neural}, which simply copies the gradients from the decoder to the encoder in model training.

Lastly, the Decoder generates a reconstructed image $I_g$ based on the quantized vectors $x_q$, which is supervised by per-pixel loss. To enhance the semantic consistency between text patches with the same index, we apply an officially trained VGG-16~\cite{simonyan2014vgg} to compute the perceptual loss~\cite{johnson2016perceptual}, where $H_g$ and $H_t$ output from the network are the high-level features of $I_g$ and $I_t$, respectively.

\begin{figure}[htb]
\centering
\small
\includegraphics[scale=.6]{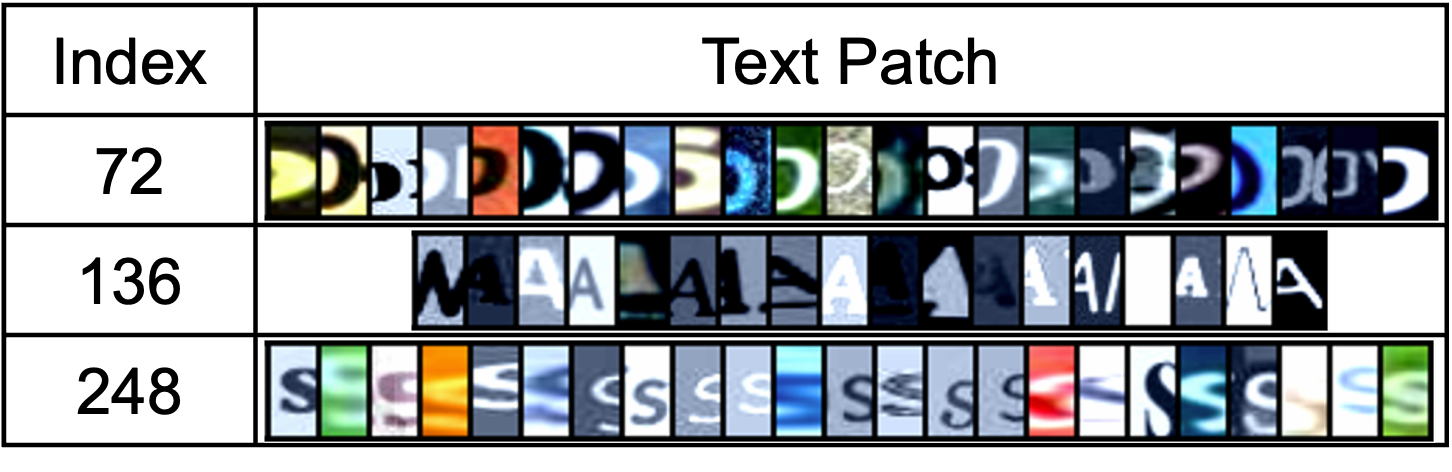}
\caption{Samples in the Text Knowledge Codebook. Three indexes correspond to parts of the letter `O', `A', and `S' respectively.}
\label{fig4}
\end{figure}

\begin{equation}
\begin{aligned}
\mathcal{L}_{_{T-VQVAE}}&=\mathcal{L}_{pixel}+\mathcal{L}_{perceptual} \\
&=\|I_g - I_t\|_2 + \|H_g - H_t\|_2~.
\end{aligned}
\end{equation}

After the T-VQVAE is well trained, text patches with similar structure and semantics can be aggregated depending on the index $z$, as shown in Fig.~\ref{fig4}. In the following SSL procedure, the frozen T-Encoder is applied to unlabelled images to serve as the Text Knowledge Codebook.

\subsection{Selective Individual Discrimination}
\label{sec:Selective Individual Discrimination}
\begin{figure*}[htb]
\centering
\small
\includegraphics[scale=.56]{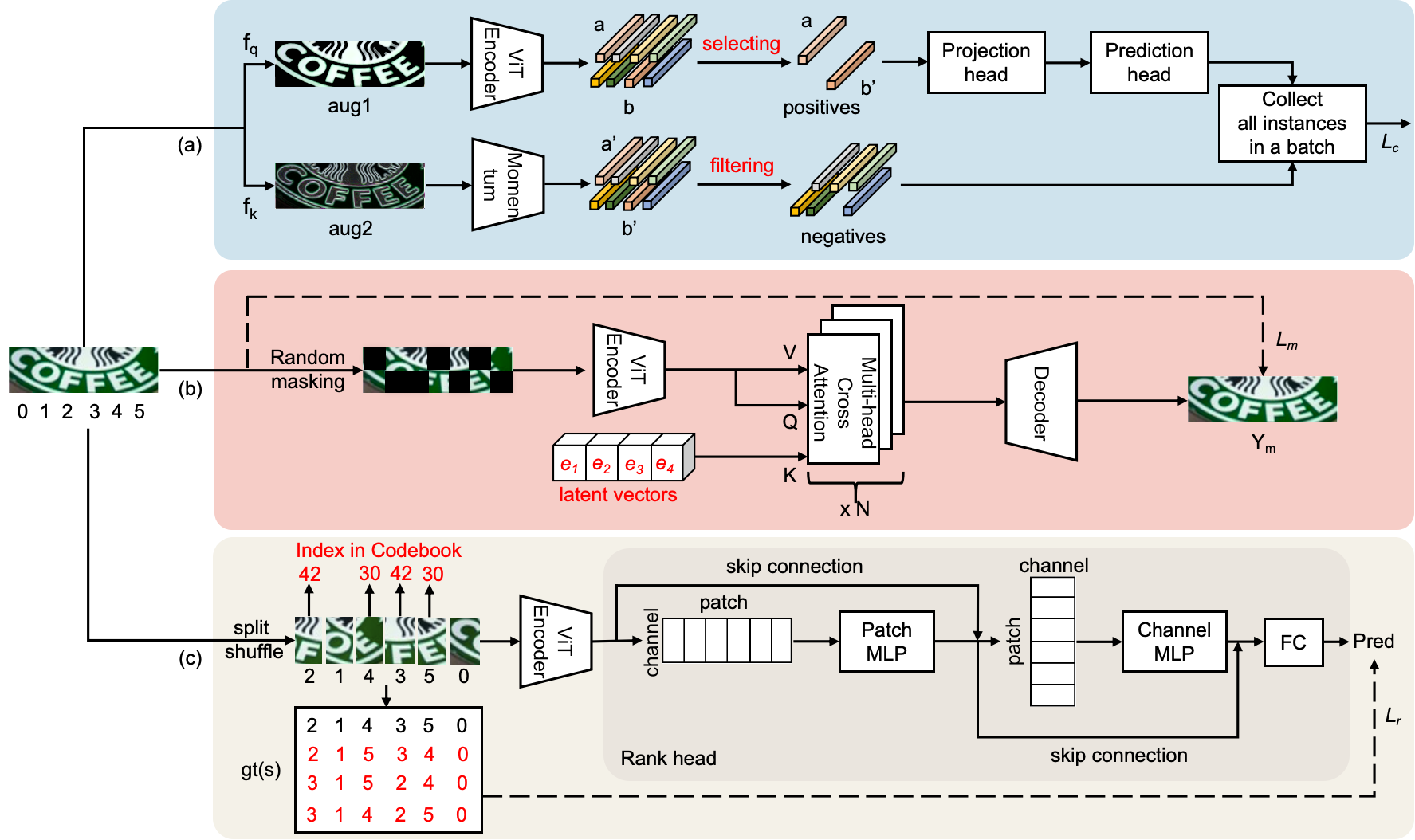}
\caption{Three SSL pretext tasks for LEGO: (a) Selective Individual Discrimination task, (b) modified Masked Image Modeling task, and (c) Random-ordered Text Rearrangement task. These three tasks are all facilitated by the \textcolor{red}{knowledge} from our Text Knowledge Codebook.}
\label{fig5}
\end{figure*}

The architecture of our discriminative task is presented in Fig.~\ref{fig5} (a), which is derived from MoCo v3~\cite{chen2021empirical}: The encoder $f_q$ consists of a ViT, a projection head, and a prediction head. The momentum encoder $f_k$ has the same components except for the projection head. In general, the input image is transformed into two different views with appropriate data augmentations. Then the anchor vectors $q$ and their corresponding vectors $k$ are obtained by $f_q$ and $f_k$, respectively. Note that the parameters of $f_k$ are updated by $f_q$ in an Exponentially Moving Average~\cite{chen2021empirical} fashion to ensure the consistency of negatives. Following the instance-mapping depicted in SeqCLR~\cite{aberdam2021sequence}, we empirically split feature maps extracted by ViT into eight horizontal patches, each of which serves as an atomic element for contrastive learning.

In addition, Selective Individual Discrimination optimizes the sample division strategy with the aid of the Text Knowledge Codebook. Different from previous contrastive learning methods, we achieve more qualitative learned representations by two schemes: \emph{filtering false negatives} and \emph{selecting positives across samples}. The T-Encoder trained in advance is exploited to quantize each input image into eight discrete indexes, thus all indexes of $q$ and $k$ can be obtained simultaneously. To perform contrastive learning more reliably, we first eliminate all false negatives (i.e. $k$ has the same index as $q$). This is because they have similar appearance and semantic information, with a high probability of belonging to the same letter or text slice. Meanwhile, we calculate the cosine similarity score between the anchor vectors $q$ and all other $k$ vectors (denoted as $k'$) in one batch that owning the same index. One of the matched pairs in top-$5$ high scores is randomly selected as a substitute positive sample. Taking advantage of this, positive samples are no longer restricted to acquisition through data augmentation, allowing us to capture more diverse positives from other images. The whole process is supervised by InfoNCE loss as $\mathcal{L}_{c}$:

\begin{equation}
\mathcal{L}_{c}=-log\frac{exp(q\cdot k'^+/\tau)}{exp(q\cdot k'^+/\tau)+\sum\limits_{i=1}exp(q\cdot k'^-/\tau)}~,
\end{equation}
where $q$ is the anchor patch, $k'^+$ and $k'^-$ are positive and negative samples through our sample division strategy.

\subsection{Modified Masked Image Modeling}
\label{sec:Modified Masked Image Modeling}
As shown in Fig.~\ref{fig5} (b), we modify the Masked Image Modeling (MIM) task to make it more suitable for perceiving the structure and location information of text. Since the patch is the fundamental processing unit of vision Transformers, it is convenient for us to adopt a patch-aligned random strategy~\cite{xie2022simmim} to manipulate masking. Following the common masking strategy, we empirically replace 75\% of the visible text patches with learnable masked tokens.

The standard MIM task treats visible patches as clues and reconstructs masked tokens to improve the representation ability of the encoder. However, the loss of entire characters or the involvement of too much background will be inevitable because of the stochastic masking strategy. This makes it difficult for the encoder to learn valid information by reconstructing these tokens. Therefore, it is necessary to introduce consistent text prior as a guide to direct the reconstruction of masked tokens with indeterminate clues. For this purpose, we retrieve corresponding latent vectors from the Text Knowledge Codebook according to indexes and incorporate them into encoded features via Multi-Head Cross-Attention blocks. Formally, we denote the latent vectors as $Q$ and masked features as $K$, and $V$. With the similarity weighting calculation in the formula~\ref{eq:5}, generalized text information in the codebook can be involved and the enhanced text features $T_e$ are obtained:

\begin{equation}
\label{eq:5}
T_e={\frac{exp(F_K\cdot F_Q^T)}{\sum_{i=1}^{n}{exp(F_K\cdot F_Q^T)}}}\cdot F_V~.
\end{equation}

Finally, we use a lightweight linear decoder to upsample $T_e$, which guarantees that there is consistent information to guide the text reconstruction regardless of the degradation of images. $\mathcal{L}_{m}$ is an $l_{1}$ loss on the masked pixels for reconstruction:

\begin{equation}
\mathcal{L}_m=\frac{1}{N(X_m)}\|Y_m-X_m\|_1~,
\end{equation}
where $X, Y\in\mathbb{R}^{3HW}$ represent the input raw images and predicted ones, respectively. The subscript ``m'' indicates all the masked patches. 

\subsection{Random-ordered Text Rearrangement}
\label{sec:Random-ordered Text Rearrangement}
The permutations of letters imply the meaning of words, which is crucial for downstream tasks such as text recognition. It is similar to humans in that they learn a new word by the arrangement of roots and affixes to understand its meaning. To this end, we design a novel Random-ordered Text Rearrangement (RTR) pretext task to implicitly model the linguistic information of text, as illustrated in Fig.~\ref{fig5} (c).

We divide the input image into $n$ portions of the same size along the horizontal axis. The initial location number of each portion is used as the ground truth of this rearrangement task. Then, we randomly disrupt their order. This task aims to predict the positional relationships between these disordered portions. Taking the sample shown in Fig.~\ref{fig5} (c) as an example, the model is required to predict the concatenation order of the rearranged image as `214350'. However, it is worth noting that if certain portions have the same content (e.g., an identical alphabet or background noise), changes in their order do not influence the meaning of words. In other words, the order relations between portions with close content can be interchanged. Accordingly, we utilize our Text Knowledge Codebook to obtain ground truth for other possible orders (i.e., `215340', `315240', and `314250' in Fig.~\ref{fig5} (c)), by comparing whether the indices corresponding to each portion in the codebook are identical.

A two-layer Mixer~\cite{tolstikhin2021mlp} is employed as rank head, which comprises alternating stacks of channel and token MLPs to fuse information obtained from different channels and spatial locations, facilitating communication support in both input dimensions. Each MLP contains two fully-connected layers and a GELU. At last, we use a classifier to predict the location to which each portion belongs.

The RTR task aims to predict the correct location for each patch. Therefore, we take the Cross-Entropy loss $\mathcal{L}_{r}$ as the optimization target:

\begin{equation}
    \mathcal{L}_r=-\sum{p \cdot log(q)}.
\end{equation}

\subsection{Optimization}

The loss function of our proposed LEGO is composed of three parts, i.e., $\mathcal{L}_{c}$, $\mathcal{L}_{m}$ and $\mathcal{L}_{r}$, enabling the concurrent optimization of three pre-training tasks (SID, MIM, and RTR):

\begin{equation}
\mathcal{L} = \mathcal{L}_{c} + \alpha\cdot\mathcal{L}_{m} + \beta\cdot\mathcal{L}_{r},
\end{equation}

where $\alpha$ and $\beta$ are scaling weights and are empirically set to $0.1$ and $1.0$, respectively. 

\section{Experiment}

\subsection{Datasets}

\textbf{Synthetic Text Data (STD)}. There are MJSynth~\cite{jaderberg2014synthetic} and SynthText~\cite{gupta2016synthetic} two synthetic datasets, containing 8.9M and 5.5M text instances respectively. We use a total of 14.4M samples for both unlabelled SSL pre-training and labelled scene text recognition downstream tasks. 

\textbf{Unlabelled Real Data (URD)}. For the sake of further exploring the potential of the pre-trained model, we use the unlabelled real dataset CC-OCR mentioned in \cite{yang2022reading} for SSL. It has about 15.77M text images, obtained by Microsoft Azure OCR system from Conceptual Captions (CC)~\cite{yang2021tap}.

\textbf{Scene Text Recognition Benchmarks}. To evaluate the recognition performance of pre-trained model, we measure the word accuracy on several real-world benchmarks which are widely used in scene text recognition studies: IIIT5K (3000)~\cite{mishra2012top}, SVT (647)~\cite{wang2011end}, IC03 (867)~\cite{lucas2005icdar}, IC13 (1015)~\cite{karatzas2013icdar}, IC15 (1811)~\cite{karatzas2015icdar}, SVT-P (645)~\cite{phan2013recognizing}, and CT80 (288)~\cite{risnumawan2014robust}.

\textbf{Scene Text Super-Resolution Benchmark}. We conduct experiments for the scene text super-resolution task to verify the adaptability of our LEGO in different downstream tasks. The pre-trained model is finetuned on 17,367 low-resolution and high-resolution training image pairs (LR-HR) of TextZoom~\cite{wang2020textzoom} dataset, which are captured by digital cameras in real scenarios. According to the focal length of digital cameras, the test set is split into easy, medium, and hard three subsets, with 1,619, 1,411, and 1,343 pairs, respectively. LR and HR images are resized to 16 × 64 and 32 × 128, respectively.

\subsection{Implementation Details}

\textbf{Pre-training}.
Data augmentation plays an important role in contrastive learning, which determines the representation quality of pre-trained models. Regarding scene text images, a variety of augmentations need to be performed to maintain the difficulty level of the discriminative task, while refraining from aggressive augmentations that result in incomplete or unreadable text. Hence, we design a proper augmentation procedure including contrast, blur, sharpen, crop, gray, color jitter, and perspective and affine transformations, from which three are randomly applied to the pre-training.

The pre-trained feature encoder is built upon a vanilla ViT-Small~\cite{dosovitskiy2020image} comprising 12 stacked Transformer blocks, where the number of heads is six and the embedding dimension $D$ is 384. The width $W$ and height $H$ of the input image are set to 128 and 32, respectively. All the pre-trained models are obtained by using four NVIDIA A100 (80GB RAM) GPUs with a batch size of 1,024. We adopt AdamW~\cite{loshchilov2018adamw} optimizer and pre-train for 20 epochs. The first epoch is warm-up, and the remaining epochs employ cosine learning rate decay, where the initial learning rate is $1.5e-4$, weight decay is 0.1, $\beta_1$ = 0.9, and $\beta_2$ = 0.95.

\begin{figure}[t]
\centering
\small
\includegraphics[scale=.35]{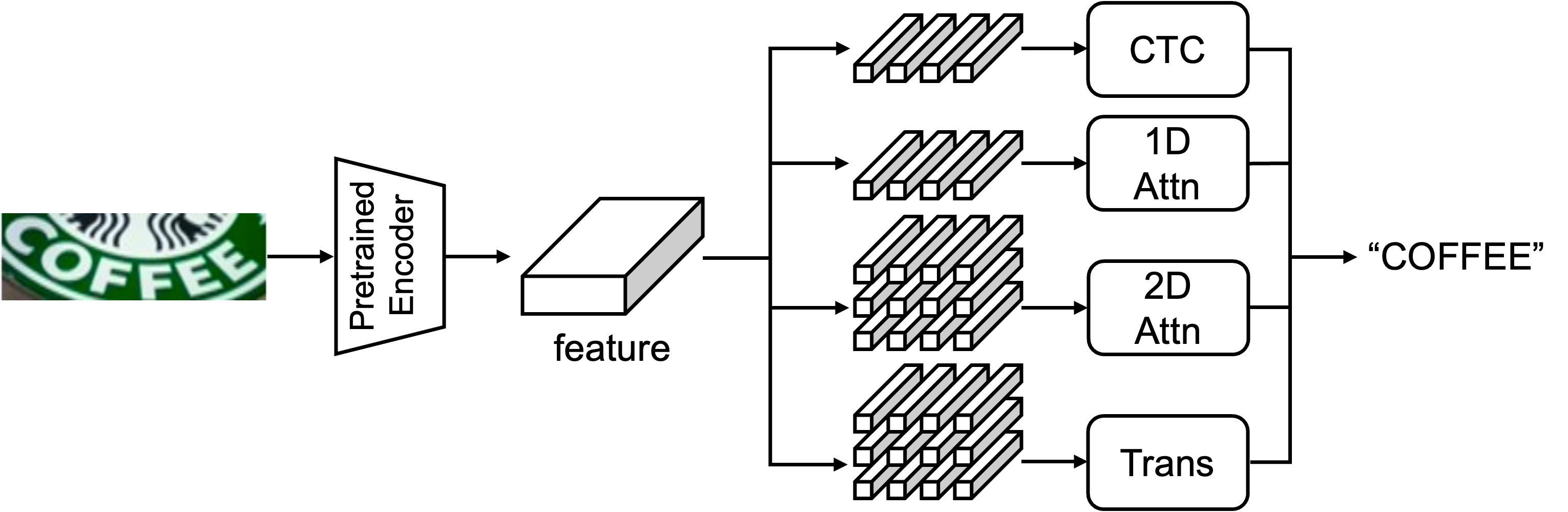}
\caption{Pipeline of text recognizer with different decoders.}
\label{fig6}
\end{figure}

\textbf{Recognition Fine-tuning}.
In general, a text recognizer consists of an encoder for extracting visual features and a decoder that converts the 2D features to a sequence of characters for prediction. Our pre-trained ViT is employed as an encoder to validate the effectiveness of the proposed method in downstream recognition tasks. Following previous methods~\cite{aberdam2021sequence, liu2022perceiving}, we inherit the same CTC~\cite{graves2006ctc, shi2016crnn} and 1D-Attention~\cite{baek2019wrong, cheng2017focusing} decoder. Furthermore, we also demonstrate the universality of our pre-trained model by using a 2D-Attention decoder in SAR~\cite{li2019sar} and a Transformer decoder in SATRN~\cite{lee2020satrn}. The complete architecture of the recognizer is illustrated in Fig.~\ref{fig6}.

We use two NVIDIA V100 (32GB RAM) GPUs to train the recognizer with a batch size of 256. AdamDelta~\cite{zeiler2012adadelta} optimizer is employed with an initial learning rate of 1 and decayed by a cosine learning rate scheduler. The entire training lasts 10 epochs, the first of which is warm-up. Common data augmentations for text images are applied randomly, including blur, contrast adjustment, noise disturbance, perspective, and affine distortion.

\textbf{Super-Resolution Fine-tuning}.
We combined the ViT encoder pre-trained by LEGO with a lightweight decoder to perform scene text super-resolution task, which is supervised by $l_2$ loss. The scale of the LR input is typically smaller than that of the HR output, which differs from the fact that the input and output of the ViT maintain the same scale. To this end, we first employ Bi-cubic interpolation to up-sample the input images to 32 × 128 before feeding them into our text super-resolution model. We fine-tune it for 200 epochs on TextZoom with a batch size of 512. Following prior works~\cite{wang2020textzoom, chen2021tbsrn, chen2021scene}, we adopt Peak Signal-to-Noise Ratio (PSNR) and Structural Similarity Index Measure (SSIM)~\cite{wang2004ssim} to evaluate the quality of reconstructed super-resolution images.

\subsection{Probe Evaluation}

Firstly, we investigate the performance of our pre-trained model with a common probe evaluation in SSL, whose classification results are positively correlated with representation quality. In concrete, we first pre-train the ViT encoder through our self-supervised LEGO on unlabelled Synthetic Text Data (STD). Then we fix the parameters of the encoder and feed the output features to the CTC or 1D-Attention decoder, which are trained on the same labeled STD.

As reported in Table~\ref{tab1}, both decoders implemented by our pre-trained encoder achieve a significantly higher word accuracy than previous approaches on three regular scene text datasets. Even though using STD only, our LEGO surpasses the PerSec~\cite{liu2022perceiving} that used 100 million extra real data for pre-training. Compared to the SimAN~\cite{luo2022siman}, we deliver at least a 3\% performance improvement regardless of which decoder is used.

In practice, we are more likely to encounter situations where there are vast amounts of unlabelled real-world data to be better exploited. Therefore, we perform self-supervised pre-training by using the URD. After adopting this new experimental setting, the recognition performance is significantly boosted, which further demonstrates the effectiveness of our approach. URD has the same data quantity as STD, but our model pre-trained by it has a more robust representation quality. We think this is because URD provides more realistic and diverse images, facilitating our pre-trained model to learn features that more closely resemble the distribution of scene text benchmarks.

\begin{table*}[htb]\small
\caption{\label{tab1}Probe evaluation on scene text recognition. `UTI-100M'' and ``Real-300K'' mean 100 million and 300 thousand extra unlabelled real images for pre-training, respectively.}
\begin{center}
\begin{tabular}{c | c | c | c | c c c}
\hline\rule{0pt}{8pt}
Method & Venue & Pre-training Data & Decoder & IIIT5K & IC03 & IC13\\
\hline\rule{0pt}{8pt}
SeqCLR~\cite{aberdam2021sequence} & CVPR'21 & SynthText~\cite{gupta2016synthetic} & \multirow{7}*{CTC} & 35.7 & 43.6 & 43.5\\
\cline{1-3}\cline{5-7}
\multirow{2}*{PerSec-ViT~\cite{liu2022perceiving}} & \multirow{2}*{AAAI'22} & STD &  & 38.4 & 46.2 & 46.7\\
 &  & STD+UTI-100M~\cite{liu2022perceiving} &  & 43.4 & 50.6 & 51.2\\
\cline{1-3}\cline{5-7}
\multirow{2}*{SimAN~\cite{luo2022siman}} & \multirow{2}*{CVPR'22} & SynthText~\cite{gupta2016synthetic} &  & 60.8 & 64.9 & 64.0\\
 &  & Real-300K~\cite{luo2022siman} &  & 68.9 & 75.0 & 72.9\\
\cline{1-3}\cline{5-7}
RCLSTR~\cite{zhang2023relational} & MM'23 & SynthText~\cite{gupta2016synthetic} &  & 54.8 & 64.8 & 60.9\\
\cline{1-3}\cline{5-7}
\multirow{2}*{LEGO-Small (Ours)} & \multirow{2}*{-} & STD &  & 64.6 & 68.8  & 67.6\\
 &  & STD+URD &  & \textbf{71.7} & \textbf{81.9} & \textbf{75.9}\\
\hline
SeqCLR~\cite{aberdam2021sequence} & CVPR'21 & SynthText~\cite{gupta2016synthetic} &  \multirow{7}*{Attention} & 49.2 & 63.9 & 59.3\\
\cline{1-3}\cline{5-7}
\multirow{2}*{PerSec-ViT~\cite{liu2022perceiving}} & \multirow{2}*{AAAI'22} & STD &  & 52.3 & 66.6 & 62.3\\
 &  & STD+UTI-100M~\cite{liu2022perceiving} &  & 55.4 & 70.9 & 66.2\\
\cline{1-3}\cline{5-7}
\multirow{2}*{SimAN~\cite{luo2022siman}} & \multirow{2}*{CVPR'22} & SynthText~\cite{gupta2016synthetic} &  & 66.5 & 71.7 & 68.7\\
 &  & Real-300K~\cite{luo2022siman} &  & 73.7 & 81.2 & 77.9\\
\cline{1-3}\cline{5-7}
RCLSTR~\cite{zhang2023relational} & MM'23 & SynthText~\cite{gupta2016synthetic} &  & 61.1 & 72.9 & 68.8\\
\cline{1-3}\cline{5-7}
\multirow{2}*{LEGO-Small (Ours)} & \multirow{2}*{-} & STD &  & 70.0 & 78.9 & 72.2\\
 &  & STD+URD &  & \textbf{79.2} & \textbf{84.8} & \textbf{82.1}\\
\hline
\end{tabular}
\end{center}
\end{table*}

\subsection{Semi-Supervision Evaluation}

We further investigate the performance under a semi-supervised manner, i.e. instead of freezing the parameters of the pre-trained ViT encoder, we fine-tune it together with the decoder. In this setting, the scene text recognizer can attain optimal performance. 

We compare the scene text recognizer implemented by our LEGO with other existing text SSL methods in Table~\ref{tab2}. To guarantee fairness, recognizers are all fine-tuned on synthetic text data. It can be seen that our proposed approach can exceed or achieve competitive recognition performance compared with existing state-of-the-art methods on both regular and irregular datasets. Concretely, when using CTC and Attention decoders, LEGO can outperform most SSL methods by relying only on STD pre-training, and the results can improve to an equivalent level with DiG~\cite{yang2022reading} after the introduction of URD. 

To further demonstrate the universality of our method, we equip our pre-trained encoder with 2D-Attention and Transformer decoder from SAR~\cite{li2019sar} and SATRN~\cite{liu2022perceiving}, respectively. Results are given in Table~\ref{tab3}.
It can be seen that compared with both original models, the model pre-trained by our LEGO achieves better recognition accuracy. At the same time, our results also exceed or are comparable with the ones pre-trained by PerSec. The excellent effectiveness and broad applicability of LEGO are further demonstrated.

\begin{table*}[ht]\small
\caption{\label{tab2}Quantitative comparisons with existing text recognition SSL methods in a semi-supervised mode. ``UTI-100M'' represents 100 million extra unlabelled real images for pre-training. All recognizers are fine-tuned on STD only.}
\begin{center}
\scalebox{0.95}{
\begin{tabular}{c | c | c | c |c c c |c c c}
\hline
Method & Venue & Pre-training Data & Decoder & IIIT5K & SVT & IC13 & IC15 & SVTP & CT80 \\
\hline
SeqCLR~\cite{aberdam2021sequence} & CVPR'21 & SynthText~\cite{gupta2016synthetic} & \multirow{6}*{CTC} & 80.9 & - & 86.3 & - & - & -\\
\cline{1-3}\cline{5-10}
\multirow{2}*{PerSec-ViT~\cite{liu2022perceiving}} & \multirow{2}*{AAAI'22} & STD &  & 83.7 & 83.0 & 89.7 & 62.3 & 70.4 & 63.5\\
 &  & STD+UTI-100M~\cite{liu2022perceiving} &  & 85.4 & 86.1 & 92.8 & 70.3 & 73.9 & 69.2\\
\cline{1-3}\cline{5-10}
DiG-Small~\cite{yang2022reading} & MM'22 & STD+URD &  & \underline{95.5} & 91.8 & \textbf{95.0} & \underline{84.1} & \underline{83.9} & \textbf{86.5}\\
\cline{1-3}\cline{5-10}
\multirow{2}*{LEGO-Small (Ours)} & \multirow{2}*{-} & STD &  & 93.4 & 89.3 & 93.7 & 81.2 & 81.6 & 84.7\\
 &  & STD+URD &  & \textbf{95.6} & \textbf{92.2} & \underline{94.8} & \textbf{84.3} & \textbf{84.7} & \underline{85.7}\\
\hline
SeqCLR~\cite{aberdam2021sequence} & CVPR'21 & SynthText~\cite{gupta2016synthetic} & \multirow{7}*{Attention} & 82.9 & - & 87.9 & - & - & -\\
\cline{1-3}\cline{5-10}
SimAN~\cite{luo2022siman} & CVPR'22 & SynthText~\cite{gupta2016synthetic} &  & 87.5 & - & 89.9 & - & - & -\\
\cline{1-3}\cline{5-10}
\multirow{2}*{PerSec-ViT~\cite{liu2022perceiving}} & AAAI'22 & STD &  & 85.2 & 84.9 & 89.2 & 70.9 & 75.9 & 69.1\\
 & AAAI'22 & STD+UTI-100M~\cite{liu2022perceiving} &  & 88.1 & 86.8 & 94.2 & 73.6 & 77.7 & 72.7\\
\cline{1-3}\cline{5-10}
DiG-Small~\cite{yang2022reading} & MM'22 & URD &  & \underline{96.4} & 94.6 & \textbf{96.6} & \textbf{86.0} & 89.3 & \textbf{88.9}\\
\cline{1-3}\cline{5-10}
\multirow{2}*{LEGO-Small (Ours)} & \multirow{2}*{-} & - &  & \textbf{93.8} & \textbf{92.3} & \textbf{94.5} & \textbf{82.3} & \textbf{85.3} & \textbf{85.9}\\
 &  & URD &  & \textbf{96.1} & \textbf{94.2} & \underline{96.4} & \textbf{86.2} & \textbf{88.8} & \underline{89.6}\\
\hline
\end{tabular}}
\end{center}
\end{table*}

\begin{table}[tb]
\caption{\label{tab3}Semi-supervision evaluation results on SAR and SATRN. PerSec and LEGO-Small introduce UTI-100M and URD for pre-training, respectively.}
\begin{center}
\scalebox{0.72}{
\begin{tabular}{c |c c c |c c c}
\hline
Method & IIIT5K & SVT & IC13 & IC15 & SVTP & CT80 \\
\hline
SAR~\cite{li2019sar} & 91.3 & 84.7 & 91.2 & 70.7 & 76.9 & 83.0\\
SAR+PerSec~\cite{liu2022perceiving} & 95.6 & 90.1 & 93.7 & 76.3 & 81.1 & 88.2\\
SAR+LEGO-Small (Ours) & \underline{95.7} & \underline{93.8} & \underline{95.3} & \underline{84.7} & \underline{87.9} & \textbf{89.6}\\
\hline
SATRN~\cite{lee2020satrn} & 94.7 & 92.1 & 94.2 & 82.1 & 86.4 & 87.6\\
SATRN+PerSec~\cite{liu2022perceiving} & 96.3 & \textbf{94.6} & \textbf{97.2} & 84.4 & \underline{89.5} & \underline{90.2}\\
SATRN+LEGO-Small (Ours) & \underline{96.4} & \underline{94.3} & \underline{97.0} & \underline{86.2} & 89.2 & \textbf{90.3}\\
\hline
\end{tabular}}
\end{center}
\end{table}

\begin{table}[htb]\small%
\caption{\label{tab3}Ablation experiments on three pretext tasks and Text Knowledge Codebook of LEGO. ``SID'', ``MIM'' and ``RTR'' denotes Selective Individual Discrimination, Modified Masked Image Modeling and Random-ordered Text Rearrangement respectively.}
\begin{center}
\begin{tabular}{c | c c c | c c c }
\hline
\multirow{2}*{Codebook} & \multirow{2}*{SID} & \multirow{2}*{MIM} & \multirow{2}*{RTR} & \multicolumn{3}{c}{CTC}\\
\cline{5-7}
 &  &  &  & IIIT5K & IC03 & IC13\\
\hline
$\times$ & \checkmark & $\times$ & $\times$ & 41.8 & 54.9 & 52.9 \\
\checkmark & \checkmark & $\times$ & $\times$ & 49.2 & 57.4 & 55.6 \\
\hline
$\times$ & \checkmark & \checkmark & $\times$ & 51.4 & 65.8 & 63.5 \\
\checkmark & \checkmark & \checkmark & $\times$ & 53.2 & 66.6 & 64.8 \\
\hline
$\times$ & \checkmark & \checkmark & \checkmark & 64.3 & \textbf{69.0} & 67.3 \\
\checkmark & \checkmark & \checkmark & \checkmark & \textbf{64.6} & \underline{68.8} & \textbf{67.6}\\
\hline
\end{tabular}
\end{center}
\end{table}

\subsection{Ablation Studies}

We validate the effectiveness of each pretext task and the Text Knowledge Codebook in our LEGO in this section. The experimental results listed in Table~\ref{tab3} are performed under the probe evaluation setting with the CTC decoder. To enumerate different pre-training strategies, SID, MIM, and RTR tasks are included gradually, and the comparisons with and without the Text Knowledge Codebook are given as well.

With the addition of three pretext tasks, the recognition accuracy improves gradually, which suggests that they all contribute to the feature representation of the encoder. We find that the performance of probe evaluation for three pretext tasks can further improve with the aid of the Text Knowledge Codebook. Among them, the most prominent increase takes place in the Individual Discrimination task. We think this is possible because the sample ambiguity problem does harm the effect of contrastive learning.

\subsection{Scene Text Super-Resolution}

In addition to common recognition tasks, pixel-level downstream tasks like scene text super-resolution can also achieve significant performance gains relying on our pre-trained model. Owing to the three pretext tasks of our LEGO, the ViT encoder is capable of handling the semantics of characters, the generation of structure, and the connective relation within words. The quantitative results are given in Table~\ref{tab4}. Despite the lack of a sophisticated design (we simply adopt a lightweight CNN as our decoder), our model still yields superior results to existing state-of-the-art methods in text super-resolution. 

\begin{table}[htb]\small
\caption{\label{tab4}Scene text super-resolution results on the TextZoom~\cite{wang2020textzoom}.}
\begin{center}
\scalebox{0.7}{
\begin{tabular}{c | c c c | c c c}
\hline
\multirow{2}*{Method} & \multicolumn{3}{c|}{SSIM} & \multicolumn{3}{c}{PSNR} \\
\cline{2-7}
 & Easy & Medium & Hard & Easy & Medium & Hard \\
\hline
SRCNN~\cite{dong2015srcnn} & 0.8152 & 0.6425 & 0.6833 & 23.13 & 19.57 & 19.56 \\
SRResNet & 0.8176 & 0.6324 & 0.7060 & 20.65 & 18.90 & 19.50\\
TSRN~\cite{wang2020textzoom} & 0.8562 & 0.6596 & 0.7285 & 22.95 & 19.26 & 19.76\\
TBSRN~\cite{chen2021tbsrn} & 0.8729 & 0.6455 & 0.7452 & 24.13 & 19.08 & 20.09\\
\hline
Scratch-ViT-Small~\cite{dosovitskiy2020image} & 0.8143 & 0.6288 & 0.6845 & 22.90 & 19.65 & 20.45\\
DiG-ViT-Small~\cite{yang2022reading} & 0.8613 & 0.6561 & 0.7215 & 23.98 & 19.85 & 20.57\\
LEGO-ViT-Small & \textbf{0.8794} & \textbf{0.6719} & \textbf{0.7476} & \textbf{24.67} & \textbf{20.58} & \textbf{20.99}\\
\hline
\end{tabular}}
\end{center}
\end{table}

\section{Conclusion}

In this study, we summarize several unique properties of text images: hierarchy, high information density, and sequentiality. These characteristics contribute to the subpar performance of general-purpose SSL methods on text images. To address this, we propose our LEGO method, which consists of three pretext tasks: SID, MIM, and RTR. All three tasks leverage our novel Text Knowledge Codebook. Specifically, in the SID task, it is used to filter out incorrect negative samples in contrastive learning, adapting to the hierarchical nature of text images. In the MIM task, the Text Knowledge Codebook provides additional prior information to the model for masked content generation, accommodating the high information density of text images. In the RTR task, the Text Knowledge Codebook can provide accurate ordering ground truth, eliminating potential uncertainties for model learning.

Extensive experiments conducted on scene text recognition benchmarks indicate that our comprehensive pre-training approach has superior representation quality and achieves better word accuracy under semi-supervision than previous state-of-the-art. Additionally, relying on a simple decoder, our pre-trained model can attain outstanding performance in scene text super-resolution tasks. In the future, we will explore incorporating codebook generation into an end-to-end network during pre-training to reduce the complexity of our pipeline and further improve the representation capability of the model.

\bibliographystyle{ieee_fullname}

\bibliography{refs}


\end{document}